\documentclass[conference]{IEEEtran}
\IEEEoverridecommandlockouts
\usepackage{cite}
\usepackage{amsmath,amssymb,amsfonts}
\usepackage{algorithmic}
\usepackage{graphicx}
\usepackage{textcomp}
\usepackage{xcolor}
\usepackage{url}
\usepackage{booktabs}
\def\BibTeX{{\rm B\kern-.05em{\sc i\kern-.025em b}\kern-.08em
    T\kern-.1667em\lower.7ex\hbox{E}\kern-.125emX}}
\begin{document}

\title{Multi-Modal Machine Learning Framework for Automated Seizure Detection in Laboratory Rats\\
\thanks{This effort was funded by the National Institute of Health (NIH), grant numbers NS079507, NS131903, and AG075583, as well as by the University of Kentucky College of Pharmacy Team Science Pilot Award.}
}

\author{\IEEEauthorblockN{Aaron Mullen}
\IEEEauthorblockA{\textit{Center for Applied Artificial Intelligence} \\
\textit{University of Kentucky}\\
Lexington, USA \\
Aaron.Mullen@uky.edu}
\and
\IEEEauthorblockN{Samuel E. Armstrong}
\IEEEauthorblockA{\textit{Center for Applied Artificial Intelligence} \\
\textit{University of Kentucky}\\
Lexington, USA \\
Sam.Armstrong@uky.edu}
\and
\IEEEauthorblockN{Jasmine Perdeh}
\IEEEauthorblockA{\textit{Department of Pharmaceutical Sciences} \\
\textit{University of Kentucky}\\
Lexington, USA \\
Jasmine.Perdeh@uky.edu}
\and
\IEEEauthorblockN{Björn Bauer}
\IEEEauthorblockA{\textit{Department of Pharmaceutical Sciences} \\
\textit{University of Kentucky}\\
Lexington, USA \\
Bjoern.Bauer@uky.edu}
\and
\IEEEauthorblockN{Jeffrey Talbert}
\IEEEauthorblockA{\textit{Center for Applied Artificial Intelligence} \\
\textit{University of Kentucky}\\
Lexington, USA \\
Jeff.Talbert@uky.edu}
\and
\IEEEauthorblockN{V.K. Cody Bumgardner}
\IEEEauthorblockA{\textit{Center for Applied Artificial Intelligence} \\
\textit{University of Kentucky}\\
Lexington, USA \\
Cody@uky.edu}
}

\maketitle

\begin{abstract}
A multi-modal machine learning system uses multiple unique data sources and types to improve its performance. This article proposes a system that combines results from several types of models, all of which are trained on different data signals. As an example to illustrate the efficacy of the system, an experiment is described in which multiple types of data are collected from rats suffering from seizures. This data includes electrocorticography readings, piezoelectric motion sensor data, and video recordings. Separate models are trained on each type of data, with the goal of classifying each time frame as either containing a seizure or not. After each model has generated its classification predictions, these results are combined. While each data signal works adequately on its own for prediction purposes, the significant imbalance in class labels leads to increased numbers of false positives, which can be filtered and removed by utilizing all data sources. This paper will demonstrate that, after postprocessing and combination techniques, classification accuracy is improved with this multi-modal system when compared to the performance of each individual data source. 
\end{abstract}


\section{Introduction}
Modality refers to the way in which a particular phenomenon is experienced. With humans, this frequently refers to our sensory inputs, such as sound, sight, or smell. Humans experience the world in a multi-modal way because we frequently use many of our different senses to draw conclusions about our surroundings. In the same way, a machine can be multi-modal if it uses multiple different sources of data to accomplish a particular task. The different modalities through which machines learn may not correlate exactly with the senses of humans, but they can be used in a similar way to improve the quality of conclusions that a machine learning model can generate.

\subsection{Background}

Much of the early work exploring multi-modality in machine learning was concerned with the combination of audio and visual mediums to extract speech information \cite{b1}. These early tests showed that the audio alone was sufficient for the models to perform well in most cases. It was only when the audio data was noisy that the visual component would be utilized to refine and improve the results from the audio signal. In situations where the audio stream was clear and noiseless, the video component did very little to improve results. This is likely the case in many multi-modal situations, where a single data source may be considered the “primary” input, and other data sources act as complementary improvements to refine the signal from the main source \cite{b2}. This is likely true even in the example of humans and our multi-modal senses; studies have suggested that around 80-85\% of the way humans interact with the world is through sight alone \cite{b3}. 

Multi-modal machine learning has become especially valuable in the field of biomedical informatics in recent years \cite{b4}. Sources like genomic data and specimen slide images can be used together to improve diagnostics. Much of the multi-modal work in the biomedical field has been focused on image segmentation and registration \cite{b5,b6}. A benefit of combining multiple image sources is that the similarity in structure between the data makes it easier to combine these modalities for a single deep learning model. In other cases, the separate data sources differ widely in format and structure, which presents a challenge to building multi-modal systems. 

Another primary challenge to multi-modality is the joining of results in a way that weighs each contribution appropriately \cite{b2}. In some cases, the data may be fed into a single model, which requires careful parameter tuning to adequately balance the importance of each source before training. In other cases, an ensemble method may be more appropriate, where separate models are trained on each modality and generate individual predictions \cite{b7}. Then, these predictions must be combined in a way that treats the contribution of each model appropriately. For example, different models may possess different levels of predictive power or noise, and these differences must be rectified for optimal results.

\subsection{Description}

Our system utilizes the ensemble technique to train a unique model on each data source and join the predictions through postprocessing techniques best suited to this particular use case. However, the system can be generalized and modified for different environments and data types. 

\begin{figure}[htbp]
\includegraphics[width=9cm]{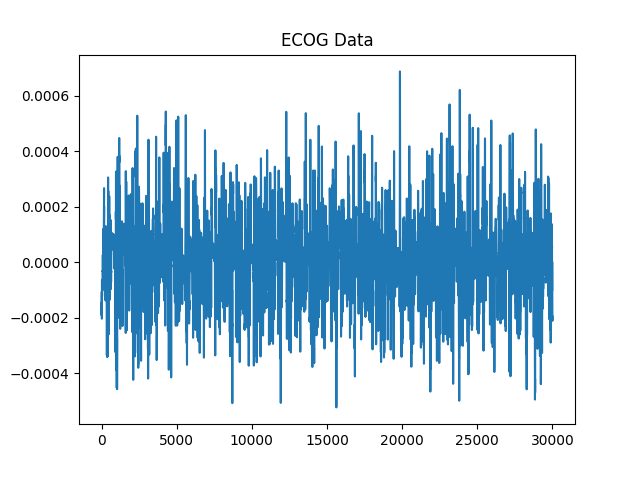}
\caption{Example of ECoG Data}
\label{ecogexample}
\end{figure}

This system was built to identify seizure events experienced in laboratory rat subjects using three different sources. The first data source is electrocorticography (ECoG) readings, which measure electrical activity in the brain. An example of this type of reading is given in Figure \ref{ecogexample}, depicting a minute long period of ECoG activity.

The ECoG data is clearly very noisy, regardless of whether a seizure is occurring or not. The transformations and processing methods used to reduce this noise and extract meaningful information from ECoG signals will be discussed further in this paper.

\begin{figure}[htbp]
\includegraphics[width=9cm]{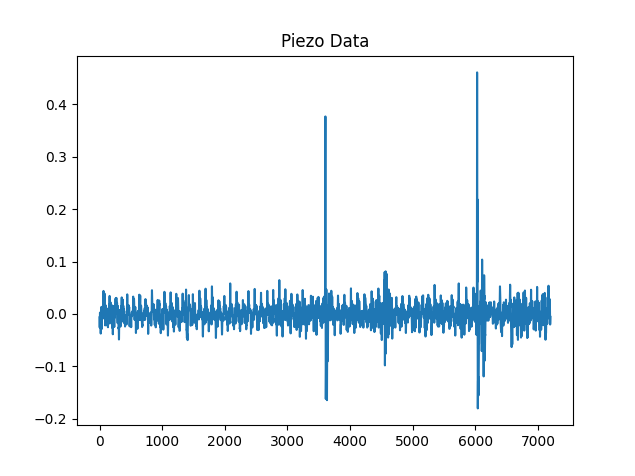}
\caption{Example of Piezo Data}
\label{piezoexample}
\end{figure}

The second source of data is piezoelectric (Piezo) motion sensor readings. An example of these readings, over the same minute-long period as depicted for the ECoG data, is given in Figure \ref{piezoexample}. This data appears less noisy, with only a few significant peaks and valleys when motion is detected.

\begin{figure}[htbp]
\includegraphics[width=9cm]{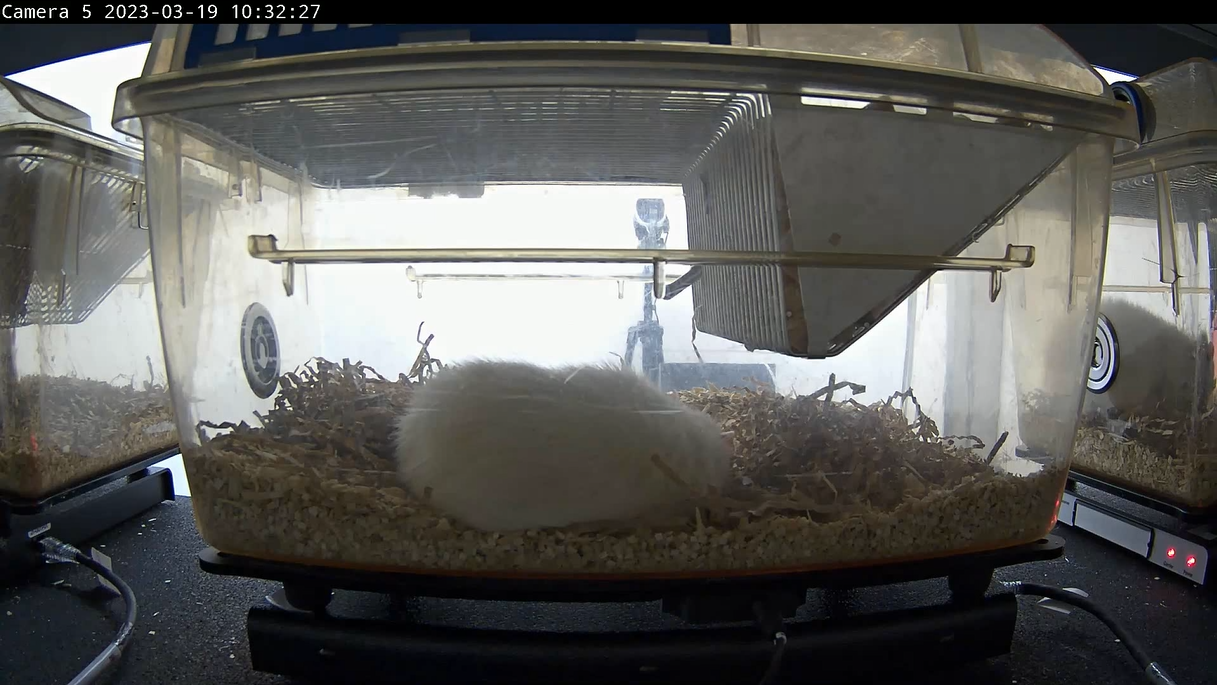}
\caption{Example Frame of Video}
\label{videoexample}
\end{figure}

The third data source utilized is recorded videos, which monitor the activity of the laboratory rats throughout the observation period. An example frame from one of these videos is shown in Figure \ref{videoexample}.

Each of these three data sources is used to train an individual model, where given time frames are classified as either containing a seizure event or not. All three models generate predictions on the same time span, and these predictions are aligned by timestamps to join the results. Finally, postprocessing techniques are used to weigh each model’s predictions appropriately, reduce noise, and generate a summary of results containing the beginning timestamp of each predicted seizure and the duration of that seizure. 

The purpose of such a system would be to automate the process of detecting seizures to limit the need for human monitoring. For example, if a research team wants to test a particular drug intended to reduce the frequency of seizures, this system can be used to monitor each subject and produce a list of experienced seizure events for each subject that can be compared and analyzed. While human intervention may be necessary to refine these results and identify false positives, the automation of the monitoring process would save hours of reviewing footage and identifying seizure events manually. 

A review of the existing literature relating to automated seizure detection finds that many models and methods have been used for different situations. Many existing solutions rely on ECoG or electroencephalogram (EEG) data, which works well in many cases but is frequently noisy and difficult to interpret \cite{b8,b9}. Other literature has acknowledged the need for multi-modal systems using multiple data sources, due to the differences in seizure types from patient to patient and high levels of noise in many existing methods \cite{b10}. Additionally, other teams have developed similar experimental designs to our own to automatically detect seizures in rat subjects, but none have implemented the same robust multi-modality \cite{b11,b12}. 

\section{Methods}

\subsection{Experimental Setup}

All animal experiments were approved by the University of Kentucky Institutional Animal Care and Use
Committee (protocol 2019-3419; PI: Bauer) and carried out per AAALAC regulations, the US Department of
Agriculture Animal Welfare Act, and the Guide for the Care and Use of Laboratory Animals of the NIH.
Female \#003 Wistar IGS (Crl:WI) rats (176 – 200 g) were obtained from Charles River Laboratories
(Kingston, NY). Rats were housed under controlled conditions (23°C, 35\% relative humidity, 12 h dark/light
cycle) with free access to tap water and standard Teklad 2018 rodent feed (Inotiv, West Lafayette, IN). To
induce status epilepticus (SE), rats received lithium chloride (127 mg/kg, i.p. injection) 16-18 hours prior to the
start of SE induction. 30 minutes prior to SE induction, rats were administered methylscopolamine (1 mg/kg,
i.p. injection). To induce SE, rats were injected with pilocarpine (30 mg/kg, i.p.) every 30 minutes until the
onset of persistent generalized seizures. The maximum pilocarpine dose per animal was limited to 160 mg/kg.
Seizures were scored based on the Racine scale \cite{b22}. Rats were classified to have SE when they had
continuous tonic-clonic seizures or intermittent class 4 or 5 seizures without recovery of normal behavior
between seizures that lasted more than 5 minutes. After 90 min of SE, rats were dosed with 10 mg/kg diazepam
by i.p. injection every 30 minutes until seizure cessation \cite{b23, b24}. Immediately following SE
resolution, rats received a subcutaneous injection with 10 mL of 5\% dextrose in normal saline for hydration.
Rats were monitored for 2 weeks to ensure complete SE resolution.

To detect spontaneous recurrent seizures 7-10 months after rats experienced SE, animals were monitored
with Adapt-a-Base piezoelectric motion-sensor platforms \cite{b25} and AXIS
M3115-LVE video cameras \cite{b26}. Motion and video data were recorded for 2
weeks with PiezoSleep® software \cite{b27}. The recorded motion data was processed
to derive a line length-based detection statistic time series, which was subjected to a threshold of 40 events per
animal per week. Each of the 40 events were reviewed on video to verify seizure occurrence. Rats with verified
spontaneous recurrent seizures were classified as chronic epilepsy (CE) rats. CE rats were implanted with HD-
S02 telemetry transmitters \cite{b28} 8-11 months after SE induction. A set of
potential leads were implanted at 7 mm posterior and 1.5 mm lateral to Bregma (positive lead) and 2 mm
anterior and 1.5 mm lateral to Bregma (negative lead). This set of potential leads was used to capture ECoG data. Animals were allowed to recover for 2 weeks after surgery before data
recording.

After the 2-week recovery period, video, ECoG, and Piezo were used to monitor CE rats. One week of
data was primarily used in the analysis for this paper. Piezo data was recorded as outlined above at a rate of 120 Hz;
video and ECoG data were recorded using PhysioTel Digital telemetry hardware, AXIS M1135 cameras, and
Ponemah® software \cite{b29}. ECoG data were sampled at 500 Hz; video was
recorded at 1920x1080 resolution at 30 fps. Ponemah® software segmented the video recordings into 30-60-minute MP4 files. Recorded ECoG data was converted to European Data Format (EDF) using NeuroScore TM \cite{b30} and raw Piezo data was converted to EDF using BIN2EDF \cite{b31}.

\subsection{Data Preprocessing}

The preprocessing and formatting of each data type will now be discussed. All coding described here was performed in Python. Because the ECoG and Piezo data follow the same overall format, these two sources will be discussed together, as they follow the same preprocessing steps. The only difference between these sources is the sampling rate at which data is collected, as there are 500 ECoG readings per second but 120 Piezo readings per second. 

Both are stored as EDF files and contain a single channel of information. This data is read in as a one-dimensional array, and timestamps are incorporated for each measurement. A separate Excel file contains recorded information for each seizure in the training data, and this file is used to assign proper class labels for each time step. 

The data is then transformed with a sliding window. Each window contains sixty seconds of data, and it slides by ten seconds for each iteration. Each window is then assigned a class label, determined by whether a recorded seizure takes place during any part of the window. This transforms the data format from a one-dimensional array to a two-dimensional structure, where each row contains a minute’s worth of measurements, and the beginning timestamp of each row is offset by ten seconds. 

\begin{figure}[htbp]
\includegraphics[width=9cm]{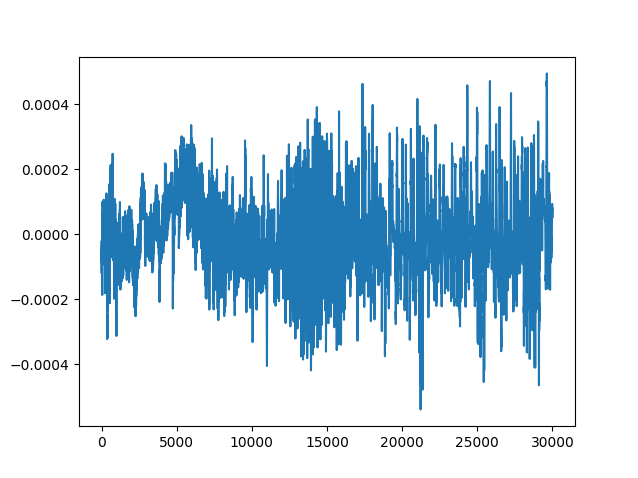}
\includegraphics[width=9cm]{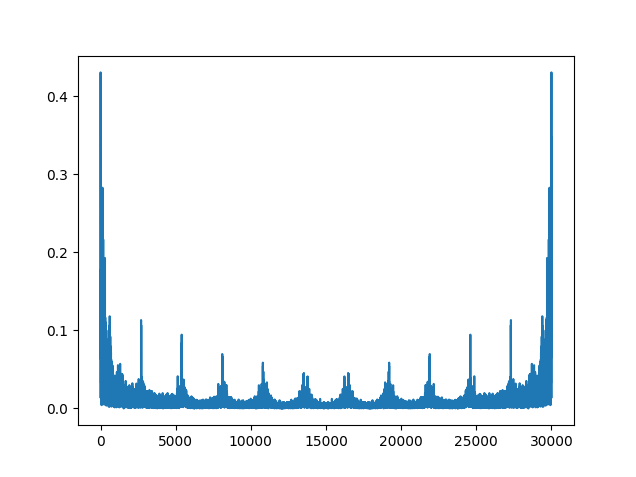}
\caption{Comparison of pre- and post-FFT.}
\label{fft}
\end{figure}

After the data is converted to this format, a Fast Fourier Transform (FFT) is performed on each window \cite{b13}, which transforms the original signal data to the frequency domain. This is done because, as stated before, ECoG signal data can be noisy and difficult to process on its own. FFT helps refine the data to focus on the most important frequency components, discarding much of the noise. It is also performed on the Piezo data for the same reasons. A comparison of a specific window of data before and after the FFT is provided in Figure \ref{fft}. The FFT is shown to transform the data dramatically, converting the signal to highlight the important frequencies.

The FFT is the last step in the preprocessing for the ECoG and Piezo data. After this, the data is passed to the models to begin training. 

Because the video data is of such a different format when compared to the other two sources, it requires completely independent methods for preprocessing. The video data was initially split up into many different files, with each video containing thirty minutes to an hour of footage. Due to the high storage and performance requirements of using a week’s worth of footage, a smaller subset of videos was used for training and testing for each animal rather than the entire dataset. Each animal was originally broken up into about 200 videos, and around 10\% of these videos were utilized. 

First, the .mp4 files are converted into .mov files for preprocessing. This was done due to the higher quality of video from less compression. Each video is then analyzed for metadata, including size and frames per second. 

Each video contains a running clock in the top left corner, as seen in Figure \ref{videoexample}, which was used to automatically acquire the starting timestamp of each video. This was done by isolating the section of the first frame of the video that contains the timestamp, normalizing it, and analyzing the image to scrape the timestamp into a string format, using the pytesseract Python library \cite{b14}, which is a Python wrapper for Tesseract-OCR, an open-source engine for character recognition in images \cite{b15}. This timestamp, along with the knowledge of frames per second, can be used to calculate the timestamp at any given frame of the video. 

Next, the starting frame of any seizures that take place within that video are identified with the timestamps, using the same reference Excel file as before. The videos are then broken up into ten-second clips, identified as either containing a seizure or not, and saved as .mp4 files. 

\subsection{Training}

After all pre-processing steps have been completed, the model training can begin. First, the training process for the ECoG and Piezo data is discussed, followed by the videos. All training and evaluation for the ECoG and Piezo data was performed on a machine equipped with an Intel Xeon v3 processor, utilizing 80 CPUs with 10 cores per socket and a clock speed of 1.9GHz. The machine has 3Tb of memory and 32GB of RAM with a speed of 2133 MHz.

The major challenge to training with the ECoG and Piezo data is the significant class imbalance. The specific class imbalance differs for each animal, but even for the animal for the most total recorded events, only 0.8\% of the timestamps contain seizures, while the remaining 99.2\% of the data falls in the negative class. Even once the sliding window steps have been implemented to bolster the count of positive seizure classifications, negative timeframes still make up over 97\% of the total dataset. Class imbalances that are far less stratified than this are known to hinder the performance of neural networks and many other classification methods \cite{b16}. We found this to be significant, as every model we trained on the full, imbalanced set performed very poorly, predicting only the majority negative class in every testing instance, even when balanced class weights were implemented. Therefore, we performed undersampling on the training data to filter out many of the negative instances and balance the dataset more evenly. This led to the question of whether to also undersample the testing set. We found that when the test set was also similarly balanced, performance was improved when compared to the imbalanced version. This is expected, as a model trained on balanced data expects balanced data for evaluation. However, balancing the test set means that the test data is no longer realistic to the real use case. When the models are used in practical situations to identify seizures during long periods, the data will not be balanced. Therefore, we decided not to perform any undersampling on the test set, leaving it imbalanced for more accurate results. As will be highlighted more in the Results section, this causes a large increase in false positives, because the model expects the test set to be similarly balanced to the training set. Therefore, multiple postprocessing techniques were implemented to reduce the number of false positives, and these will be discussed shortly. 

Two primary types of models were evaluated on the ECoG and Piezo data. One, a standard recurrent neural network (RNN) system was implemented using the Tensorflow library \cite{b17}. The RNN contains a single hidden layer and uses an imbalanced validation set, optimizing for the highest precision, recall, and AUC using the Adam optimizer \cite{b18}. Accuracy is not used due to the high imbalance.  

Secondly, a Time Series Forest model from the Pyts library \cite{b19} was utilized as well. Pyts is a library built specifically for time series classification, which is the method of classifying segments of time series data. The ECoG and Piezo data, in their windowed format, are essentially collections of minute-long time series segments, each of which can be classified separately. The Time Series Forest method is similar to a random forest classifier, fitting separate decision trees on different subsets of the time series and calculating features such as mean, standard deviation, and slope of each subset. 

Both models are trained on the data from two rat subjects and evaluated on a test set made up of the data from a third subject. These subjects were chosen because they had the most recorded seizure events. The class balance of each dataset, after the sliding window transformation was performed, is provided in Table \ref{classbalance}.

\begin{table}[htbp]
\caption{Class Imbalance of Each Subject}
\begin{center}
\begin{tabular}{lll}
\toprule
Animal         & Percent Negative Class & Percent Positive Class \\ \midrule
Train Animal 1 & 99.7\%                 & 0.3\%                  \\
Train Animal 2 & 97.7\%                 & 2.3\%                  \\
Test Animal    & 99.8\%                 & 0.2\%                  \\ \bottomrule
\end{tabular}
\label{classbalance}
\end{center}
\end{table}

Each animal is significantly imbalanced, presenting challenges to training and evaluation. During inference, each model produces predictions for each minute-long time frame in the form of a probability. For example, a value of 0.8 at a timestep indicates 80\% confidence in a positive seizure classification for that time frame. Finally, a file containing each timestamp, predicted probability, and real classification label is saved. A comparison of the performance of each model on each dataset will be discussed in the Results section. 

\begin{figure*}[htbp]
\includegraphics[width=18cm]{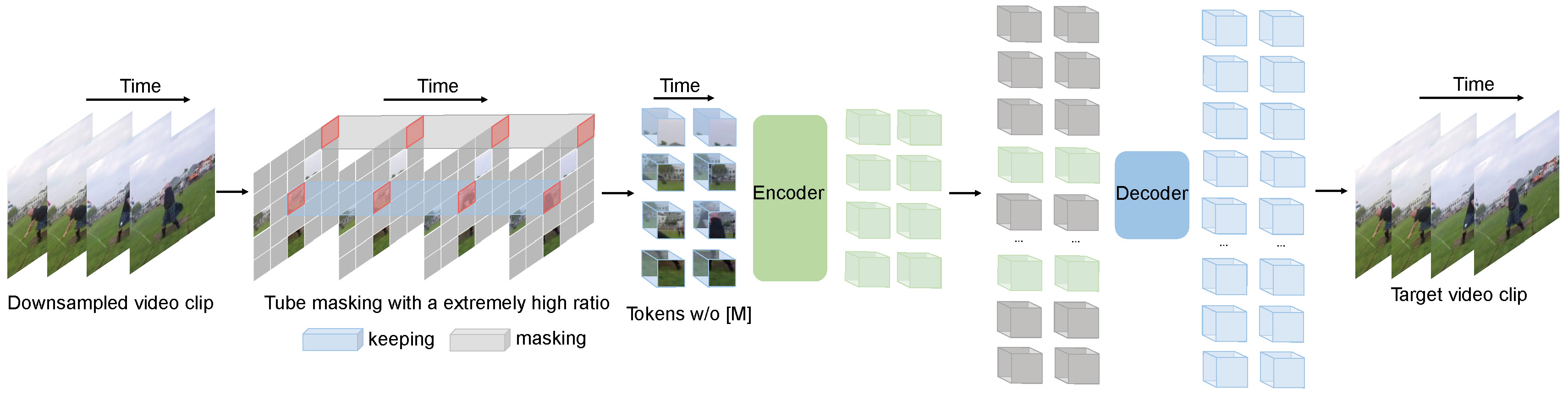}
\caption{VideoMAE Model Diagram \cite{b21}}
\label{videomae}
\end{figure*}

Similarly to the preprocessing, the videos must be handled in an entirely separate way. The videos were processed on an Nvidia Quadro RTX 5000 GPU machine. Because only subsets of the total videos were preprocessed, no sampling is necessary. The video training is performed using transformers for image processing and video classification, called VideoMAE, which were described in a publication by Tong, Song, Wang, and Wang \cite{b20}. These base models are available for use on Hugging Face \cite{b21}. A diagram summarizing the steps utilized in the VideoMAE model is provided in Figure \ref{videomae}. 

After normalizing the video frames, training is done with randomly sampled clips of two to three seconds from all videos in the training set. Similarly to the ECoG and Piezo process, the model is trained on data from two animals and tested on a third. The VideoMAE model uses a validation set during training and produces predictions on similarly sampled clips from the test set. The output of the evaluation does not produce classification probabilities as with the ECoG and Piezo data, so only the discrete predicted classification values are used here.

\subsection{Postprocessing}

After all models have generated their predictions, postprocessing is necessary to clean the results and combine each model’s output. As stated before, the method of training a model on a balanced dataset but testing on an imbalanced one leads to many false positive predictions that need to be filtered out. One simple way to improve the initial accuracy of the output is to filter out any isolated positive predictions. The seizures recorded in this dataset tend to last between thirty and sixty seconds. Since the sliding window only moves by ten seconds for each iteration, a positive seizure event will last for several iterations. Therefore, a positive prediction for a single timeframe that is surrounded by only negative predictions is unlikely to be correct. Therefore, all isolated positive predictions are filtered out. 

Additionally, the inclusion of classification probabilities for the ECoG and Piezo results allows for tweaking of the threshold value for positive predictions. By increasing this threshold, many false positives can be filtered out, but at the risk of possibly missing real seizure events that the model was simply less confident in. This is where a multi-modal system is advantageous, as a single model’s output would struggle with this problem, while a multi-modal system can combine results from different models to better determine what should be considered a valid positive prediction based on how often those models’ outputs agree. 

Finally, the output of all three models is combined, using different weights for each system to maximize performance. The final list of predictions is processed once more, grouping runs of positive predictions together. A final document is generated that includes beginning timestamps for each predicted seizure, the duration of that seizure, and how confident the model is in that prediction, using the predicted probabilities.

\section{Results}

\subsection{Performance Metrics}

First, a discussion of the performance metrics used here is necessary. Individually, we produced metrics for accuracy, AUC, precision, and recall. Due to the high number of false positives, precision is usually quite low, and accuracy is not as useful as other metrics for heavily imbalanced datasets. Therefore, AUC and recall are the most important metrics for comparing these results. 

After the results have been combined and processed, different performance metrics are useful, as a direct comparison of positive and negative class predictions for each timeframe can be misleading. For example, a seizure event may begin at timestamp 4 and last until timestamp 8. However, the predicted seizure may begin at timestamp 1 and end at timestamp 10. If standard accuracy was calculated, this would count against the model’s performance. Clearly, in this example, the model identified the seizure accurately, but due to the way the sliding window is constructed, this positive prediction began early and ended late. Therefore, for the final results, both standard accuracy metrics and custom metrics are given to provide a full picture of performance. Custom metrics will calculate how many seizure events were correctly identified, based on any overlap in the timeframe between real seizures and predicted seizures, and how many false positive seizure events were generated as well. 

\subsection{Individual ECoG Results}

To demonstrate the impact of postprocessing and the combination of results, the individual results for each data source will be discussed first. For each of these cases, isolated positive predictions were removed, but no other processing or combination was performed. First, the ECoG results are given in Table \ref{individualeeg}, for both the RNN model and the time series forest model. 

\begin{table}[htbp]
\caption{Individual Results for ECoG Data}
\begin{center}
\begin{tabular}{lll}
\toprule
Performance Metric & RNN  & Time Series Forest \\ \midrule
Accuracy           & 0.69 & 0.96               \\
AUC                & 0.76 & 0.87               \\
Precision          & 0.0  & 0.03               \\
Recall             & 0.82 & 0.77               \\ \bottomrule
\end{tabular}
\label{individualeeg}
\end{center}
\end{table}

Table \ref{individualeeg} shows that the time series forest model performs better overall, but both models produce results with high numbers of false positives, as shown in the low precision. Still, the time series forest model performed better in this regard, misclassifying 2,700 total time steps as containing seizures, compared to the 19,000 timesteps that the RNN incorrectly identified as positive. However, the RNN has better recall, correctly classifying more positive seizure instances as positive. 

The results from the time series forest model will be used as the primary results. It is useful to frame these results more practically by explaining how many seizures were successfully detected and how many false positives were identified. The specific results depend on the threshold value chosen, where higher thresholds reduce false positives but also prevent the model from catching some real seizure events. Out of the ten total seizures in the test set, the highest threshold value that identifies all ten also detects 162 false positive events.  

To justify the use of FFT, results are provided in Table \ref{individualeegnofft} from a time series forest model trained and tested on data that had not undergone FFT. 

\begin{table}[htbp]
\caption{Individual Results for ECoG Data Without FFT}
\begin{center}
\begin{tabular}{ll}
\toprule
Performance Metric & Time Series Forest (No FFT) \\ \midrule
Accuracy           & 1.0                         \\
AUC                & 0.5                         \\
Precision          & 0.0                         \\
Recall             & 0.0                         \\ \bottomrule
\end{tabular}
\label{individualeegnofft}
\end{center}
\end{table}

This model solely predicts the negative class, leading to almost perfect accuracy but no precision or recall. It was trained on the same dataset size as the regular ECoG model, proving that FFT is a valuable step in preprocessing. 
For further comparison, results are provided for an evenly balanced test set in Table \ref{individualeegbalanced}. 

\begin{table}[htbp]
\caption{Individual Results for ECoG Data on Balanced Test Set}
\begin{center}
\begin{tabular}{ll}
\toprule
Performance Metric & Time Series Forest (Balanced Test Set) \\ \midrule
Accuracy           & 0.89                                   \\
AUC                & 0.89                                   \\
Precision          & 1.0                                    \\
Recall             & 0.77                                   \\ \bottomrule
\end{tabular}
\label{individualeegbalanced}
\end{center}
\end{table}

These results are similar to the performance on the unbalanced test set in recall and AUC but with slightly lower accuracy in exchange for much higher precision. The model can perform very well on ECoG data alone when given an even balance of positive and negative classes; however, due to the inherent imbalance in the real data, other data sources are necessary to bolster this performance. 

\subsection{Individual Piezo Results}

The standard, individual results for the Piezo data are provided in Table \ref{individualpiezo}. 

\begin{table}[htbp]
\caption{Individual Results for Piezo Data}
\begin{center}
\begin{tabular}{lll}
\toprule
Performance Metric & RNN  & Time Series Forest \\ \midrule
Accuracy           & 0.06 & 0.58               \\
AUC                & 0.53 & 0.73               \\
Precision          & 0.0  & 0.0                \\
Recall             & 0.99 & 0.88               \\ \bottomrule
\end{tabular}
\label{individualpiezo}
\end{center}
\end{table}

The results from the models trained solely on the Piezo data source are much worse than those from the ECoG models. In this instance, the RNN classifies most timesteps as positive, achieving nearly perfect recall but suffering in every other metric. Again, the time series forest provides more balanced results. Still, precision is incredibly low, and when analyzed, 1,002 total false positive events are detected at the best threshold value. 

\subsection{Individual Video Results}

The individual results for the video classifier are provided below, trained using the VideoMAE model. After the results were reorganized into chronological order, isolated positive predictions were removed, and the performance metrics are given in Table \ref{individualvideo}. 

\begin{table}[htbp]
\caption{Individual Results for Video Data}
\begin{center}
\begin{tabular}{ll}
\toprule
Performance Metric & VideoMAE \\ \midrule
Accuracy           & 0.78     \\
AUC                & 0.78     \\
Precision          & 0.72     \\
Recall             & 0.91     \\ \bottomrule
\end{tabular}
\label{individualvideo}
\end{center}
\end{table}

91\% of clips that contained a seizure were correctly identified, while 65\% of clips not containing a seizure were correctly identified. This shows that false positives are common with the video classification as well. The precision is higher here when compared to the ECoG and Piezo results, but it is important to note that the subset of videos that this model was tested on is more balanced than the entire dataset. When the predictions for each sampled clip of each time step are averaged together and evaluated, all ten correct seizures are identified, with only a single false positive. 

\subsection{Combined Results}

First, the ECoG and Piezo results are combined, as these results both cover the entire test set. The results from the time series forest, which performed better individually, are used. Standard accuracy metrics are given in Table \ref{combinedeegpiezo}.

\begin{table}[htbp]
\caption{Combined Results for ECoG and Piezo Data}
\begin{center}
\begin{tabular}{ll}
\toprule
Performance Metric & Combined (ECoG and Piezo) \\ \midrule
Accuracy           & 0.96                     \\
AUC                & 0.89                     \\
Precision          & 0.03                     \\
Recall             & 0.82                     \\ \bottomrule
\end{tabular}
\label{combinedeegpiezo}
\end{center}
\end{table}

These results are very similar to the results from the ECoG data alone. This is because the Piezo data contains more false positives, and therefore, its contribution to the overall prediction is weighted less. However, recall and AUC improve slightly while other metrics remain consistent. 

The highest threshold level that catches all ten seizures correctly also identifies 108 false positive seizure events. This is a significant improvement over the number of false positives identified in both the ECoG and Piezo individual results, but it is still low in precision. The results can be improved further by incorporating the video element. When all three data sources are evaluated together, standard performance metrics are given in Table \ref{combined}. 

\begin{table}[htbp]
\caption{Combined Results for ECoG, Piezo, and Video Data}
\begin{center}
\begin{tabular}{ll}
\toprule
Performance Metric & Combined (ECoG, Piezo, Video) \\ \midrule
Accuracy           & 0.99                         \\
AUC                & 0.89                         \\
Precision          & 0.17                         \\
Recall             & 0.82                         \\ \bottomrule
\end{tabular}
\label{combined}
\end{center}
\end{table}

These results are, again, similar to the others, but with improved precision. When analyzing the number of detected seizures, the threshold can be increased, as the inclusion of video data allows the ensemble to be more confident in positive predictions. The highest threshold level that correctly identifies all ten positive seizure events now identifies 32 false positives, which is an improvement over the 108 identified without the inclusion of video data. An overall comparison of the number of false positives is given in Table \ref{falsepositive}.

\begin{table}[htbp]
\caption{False Positive Results}
\begin{center}
\begin{tabular}{ll}
\toprule
Model             & Number of False Positives \\ \midrule
ECoG               & 162                       \\
Piezo             & 1002                      \\
ECoG, Piezo        & 108                       \\
ECoG, Piezo, Video & 32                        \\ \bottomrule
\end{tabular}
\label{falsepositive}
\end{center}
\end{table}

The value for false positives for the video model alone is not provided, as the video model was individually trained and tested on a smaller subset of data. This is also relevant to the results when all three data sources are combined. Because the video model only generated results for a smaller subset of data, not every timestep has a prediction from all three data sources. Therefore, timeframes that did not contain video predictions simply used the predictions from ECoG and Piezo data alone, while timeframes that did contain video predictions used a weighted average of all three data sources. Likely, this has positively skewed the precision of the combined results, as there are fewer opportunities for false positives to be generated by the video model. Future work will be done to incorporate more video subsets, but we are confident that the video model would still improve overall results.

\subsection{Inference Times}

A brief discussion of tested inference times is included here. A potential usage of this system would be to perform real-time analysis using live data streams containing ECoG, Piezo, and video signals. This system could identify potential seizure events in real-time using a combination of all three sources. A comparison of the inference times for each of the three models, using the Time Series Forest model for ECoG and Piezo and the VideoMAE model for the videos, is given in Table \ref{inf}. These values represent the time, in seconds, that it takes for a ten second segment of data to be fully processed, on the same machines as highlighted in the Methods section.

\begin{table}[htbp]
\caption{Inference Times (per ten seconds)}
\begin{center}
\begin{tabular}{ll}
\toprule
Data Source & Inference Time (s) \\ \midrule
ECoG        & 1.76               \\
Piezo       & 0.10               \\
Video       & 5.68               \\ \bottomrule
\end{tabular}
\label{inf}
\end{center}
\end{table}

The video value of 5.68 represents that it takes 5.68 seconds for a 10-second clip to be fully processed and classified. The values for the ECoG and Piezo data are for a minute-long segment, but because the sliding window moves by ten seconds for each iteration, the results are comparable. These segments are able to be processed faster than the videos, particularly the Piezo data. The combined requirement of 7.64 seconds of processing for every ten seconds of data indicates that real-time processing and classification would be possible.

\section{Discussion}

The results demonstrate that the inclusion of multiple data sources improved upon original, individual results in every case. Individually, both ECoG and Piezo performed well in recall and AUC, but particularly poorly in precision. Performance was high when both the training and testing data were balanced, but in the interest of realism, both models were trained on balanced data and evaluated on imbalanced data, generating a high number of false positives. Many of these false positives can be filtered out by removing isolated positive predictions and increasing threshold values, but even after these steps, the count of false positives always outnumbers the count of true seizures. 

The time series forest model outperformed the RNN model for both data sources. However, the RNN does possess several advantages, including faster training and evaluation time and smaller model storage requirements. It is possible that more parameter tuning could improve the RNN’s performance in future work. Still, the time series forest model, structured specifically for this kind of time series classification problem, was used as the primary model for these data sources. 

The test set used contained only ten positive seizure events throughout over 60,000 distinct time frames. This level of high imbalance naturally makes any prediction task difficult. Certain anomaly detection strategies exist for this kind of imbalanced data, and these new models may be implemented in the future. However, they were not included at this time, primarily because most anomaly detection methods focus on identifying individual data points that fall outside expected ranges, but these do not work as well for classifying entire time series segments. 

The video model was evaluated on a more balanced test set, due to the high storage and performance cost of using the entire video dataset. Relatedly, the video individual results are the most balanced of any of the data sources. It is likely that, if evaluated on the entire test set, the number of false positives would increase significantly. However, we are confident that, after postprocessing, the quality of the video results would still hold up well, for several reasons. Due to the random clip sampler, there are several predictions made for each time frame, and the overall prediction from that time frame is determined based on the majority classification, which allows for some false positives to be removed. Additionally, after combining results with the other data sources, the probability of all three models incorrectly classifying a certain time frame is expected to be much lower than when only two, very similar models are used from the ECoG and Piezo data. 

Even still, the combination of just ECoG and Piezo results leads to better performance than when evaluated individually. The Piezo individual performance is quite poor, but it serves as a useful complement to the stronger, primary ECoG data. Because predicted probabilities are used rather than discrete classifications, the most confident Piezo predictions can be used to fill in gaps in the ECoG’s classifications without the abundance of less confident false positives causing a detriment to performance. The incorporation of the video model improves performance further by increasing confidence in positive predictions. In the future, work will be done to evaluate more video samples to better assess the impact of the VideoMAE model on the overall results. 

While the final results still show more false positives than correct predictions, we believe that this system still demonstrates a very useful technique for identifying seizure events. Given that each predicted seizure lasts between thirty seconds and a minute, it would likely take a researcher less than an hour to review footage of all predicted seizures to manually remove false positives and produce a final, accurate count of seizure events that occurred. This is a massive improvement over having to manually review hundreds of hours of footage to make these identifications. Therefore, this system represents a valuable tool in automating the seizure detection process.

\section{Conclusion}

This article has presented a system for multi-modal time series classification, which has been demonstrated using an example of automated seizure detection in laboratory rat subjects. Three data sources were utilized, including ECoG, Piezo, and video. Both ECoG and Piezo data were handled using time series forest models, while the videos were trained with the VideoMAE architecture. Due to the high imbalance in positive vs. negative classifications, each model struggled with false positives, but by combining all models’ output, the true positive classifications can be bolstered while many of each model’s false positives are filtered out. This leads to improved performance when compared to all individual models. Additionally, the inference times are low enough to demonstrate the capability of real-time classification. This system proves the value of multi-modality, particularly with imbalanced datasets, and demonstrates its usefulness in cleaning messy results while maintaining confident predictions. 



\end{document}